\documentclass[11pt,a4paper]{article}
\usepackage[hyperref]{acl2021}
\usepackage{times}
\usepackage{latexsym}

\usepackage{kotex}
\usepackage{amssymb}
\usepackage{amsmath}
\usepackage{dsfont}
\usepackage{makecell}
\usepackage{multirow}
\usepackage{graphicx}
\usepackage{subfigure}
\usepackage[normalem]{ulem}

\usepackage{microtype}

\aclfinalcopy 
\def\aclpaperid{2373} 

\title{Attend What You Need: \\Motion-Appearance Synergistic Networks for Video Question Answering}

\author{Ahjeong Seo$^{1}$, Gi-Cheon Kang$^{1, 2}$, Joonhan Park$^{3,}$\thanks{ \;\; Work done during an internship at AI Institute for Seoul National University (AIIS).} , Byoung-Tak Zhang$^{1,2}$ \\
$^{1}$Seoul National University \\
$^{2}$AI Institute for Seoul National University (AIIS) \\
$^{3}$ Hanyang University \\
\texttt{\{ajseo,gckang,jhpark,btzhang\}@bi.snu.ac.kr} \\ 
}

\date{}

\begin{document}

\maketitle
\begin{abstract}

Video Question Answering is a task which requires an AI agent to answer questions grounded in video. This task entails three key challenges: (1) understand the intention of various questions, (2) capturing various elements of the input video (\textit{e.g.,} object, action, causality), and (3) cross-modal grounding between language and vision information. We propose \textbf{M}otion-\textbf{A}ppearance \textbf{S}ynergistic \textbf{N}etworks (MASN), which embed two cross-modal features grounded on motion and appearance information and selectively utilize them depending on the question's intentions. 
MASN consists of a motion module, an appearance module, and a motion-appearance fusion module. The motion module computes the action-oriented cross-modal joint representations, while the appearance module focuses on the appearance aspect of the input video. Finally, the motion-appearance fusion module takes each output of the motion module and the appearance module as input, and performs question-guided fusion. As a result, MASN achieves new state-of-the-art performance on the TGIF-QA and MSVD-QA datasets. We also conduct qualitative analysis by visualizing the inference results of MASN. The code is available at \url{https://github.com/ahjeongseo/MASN-pytorch}.

\end{abstract}

\begin{figure*}[t]
\centering
    \includegraphics[width=16cm]{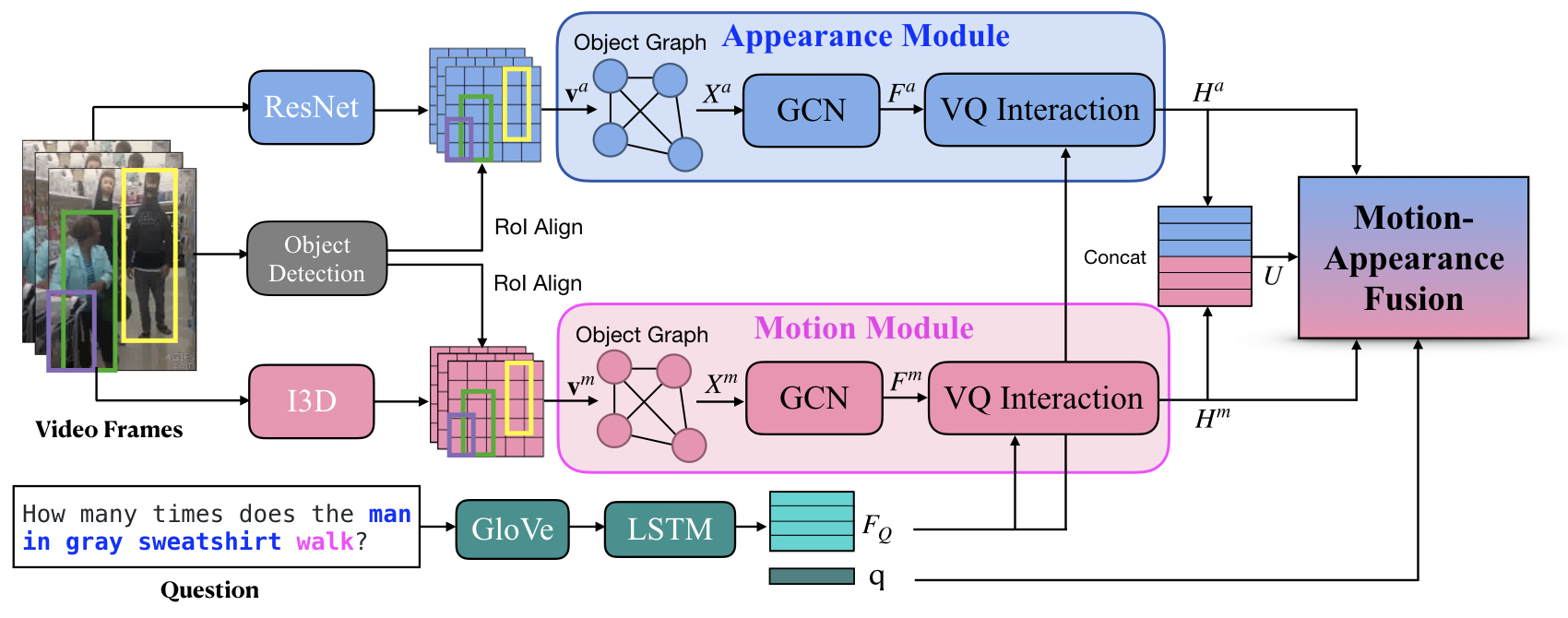}
\caption{An overview of MASN. Each extracted feature from ResNet and I3D is fed into the Appearance and Motion modules. Both modules have the same structure with a GCN and VQ interaction submodule. The results from each module are then concatenated and fused in the Motion-Appearance Fusion module. The output from the fusion module is used to derive answers. For question features, the word-level representation $F_{Q}$ is integrated with the visual features in the VQ interaction submodule. The last hidden units $\mathbf{q}$ from the bi-LSTM are used to combine appearance and motion features.}
\label{fig:model1}
\end{figure*}

\section{Introduction}
Recently, research in natural language processing and computer vision has made significant progress in artificial intelligence (AI). Thanks to this, vision-language tasks such as image captioning \cite{xu2015show}, visual question answering (VQA) \cite{antol2015vqa,goyal2017making}, and visual commonsense reasoning (VCR) \cite{zellers2019recognition} have been introduced to the research community, along with some benchmark datasets. In particular, video question answering (video QA) tasks \cite{xu2016msr,jang2017tgif,lei2018tvqa,yu2019activitynet,choi2020dramaqa} have been proposed with the goal of reasoning over higher-level vision-language interactions. In contrast to QA tasks based on static images, the questions presented in the video QA dataset vary from frame-level questions regarding the appearance of objects (\textit{e.g.,} what is the color of the hat?) to questions regarding action and causality (\textit{e.g.,} what does the man do after opening a door?). 


There are three crucial challenges in video QA: (1) understand the intention of various questions, (2) capturing various elements of the input video (\textit{e.g.,} object, action, and causality), and (3) cross-modal grounding between language and vision information. To tackle these challenges, previous studies \cite{li2019beyond, jiang2020divide, huang2020location} have mainly explored this task by jointly embedding the features from the pre-trained word embedding model \cite{pennington2014glove} and the object detection models \cite{he2016deep,ren2016faster}. However, as discussed in \cite{gao2018motion}, the use of the visual features extracted from the object detection models suffers from motion analysis since the object detection model lacks temporal modeling. To enforce the motion analysis, a few approaches \cite{xu2017video,gao2018motion} have employed additional visual features \cite{tran2015learning} (\textit{i.e.,} motion features) which were widely used in the action recognition domain, but their reasoning capability is still limited. They typically employed recurrent models (\textit{e.g.,} LSTM) to embed a long sequence of the visual features. Due to the problem of long-term dependency in recurrent models \cite{bengio1993problem}, their proposed methods may fail to learn dependencies between distant features.

In this paper, we propose Motion-Appearance Synergistic Networks (MASN) for video question answering which consist of three kinds of modules: the motion module, the appearance module, and the motion-appearance fusion module. As shown in Figure \ref{fig:model1}, the motion module and the appearance module aim to embed rich cross-modal representations. These two modules have the same architecture except that the motion module takes the motion features extracted from I3D as visual features and the appearance module utilizes the appearance features extracted from ResNet. Each of these modules first constructs the object graphs via graph convolutional networks (GCN) to compute the relationships among objects in each visual feature. Then, the vision-question interaction module performs cross-modal grounding between the output of the GCNs and the question features. The motion module and the appearance module each yield cross-modal representations of the motion and the appearance aspects of the input video respectively. The motion-appearance fusion module finally integrates these two features based on the question features. 

The main contributions of our paper are as follows. First, we propose Motion-Appearance Synergistic Networks (MASN) for video question answering based on three modules, the motion module, the appearance module, and the motion-appearance fusion module. Second, we validate MASN on the large-scale video question answering datasets TGIF-QA, MSVD-QA, and MSRVTT-QA. MASN achieves the new state-of-the-art performance on TGIF-QA and MSVD-QA. We perform ablation studies to validate the effectiveness of our proposed methods. Finally, we conduct a qualitative analysis of MASN by visualizing inference results.

\section{Related Work}
\textbf{Visual Question Answering (VQA)} is a task that requires both understanding questions and finding clues from visual information. VQA can be classified into two categories based on the type of the visual source: image QA and video QA. In image QA, earlier works approach the task by applying attention between the question and the spatial dimensions of the image \cite{yang2016stacked, anderson2018bottom, kim2018bilinear, kang2019dual}. In video QA, since a video is represented as a sequence of images over time, recognizing the movement of objects or causality in the temporal dimension should also be considered along with the details from the spatial dimension \cite{jang2017tgif, on2020cut}. There have been some attempts \cite{xu2017video, gao2018motion, fan2019heterogeneous} to extract motion and appearance features and integrate them on a spatio-temporal dimension via memory networks. \citet{li2019beyond}, \citet{huang2020location}, \citet{jiang2020divide} proposed better performing models using attention in order to overcome the long-range dependency problem in memory networks. However, they do not represent motion information sufficiently since they only use features pre-trained on image or object classification. To better address this, we model spatio-temporal reasoning on multiple visual information (\textit{i.e.,} ResNet, I3D) while also solving the long-range dependency problem that occurred in previous studies. 

\noindent\textbf{Action Classification} is a task of recognizing actions, which are composed of interactions between actors and objects. Therefore, this task has much in common with video QA, in that the model should perform spatio-temporal reasoning. For better spatio-temporal reasoning, \citet{tran2015learning} introduced C3D, which extends the 2D CNN filters to the temporal dimension. \citet{carreira2017quo} proposed I3D, which integrates 3D convolutions into a state-of-the-art 2D CNN architecture, which now acts as a baseline in action classification tasks \cite{murray2012ava, girdhar2018better}. \citet{feichtenhofer2019slowfast} introduced SlowFast, a network which encodes images in two streams with different frame rates and temporal resolutions of convolution. This study based on a two-stream architecture inspired us in terms of assigning different inputs to each encoder module. However, our method differs from the former studies in two aspects: (1) we utilize language features as well as vision features, and (2) we expand the two-stream structure to solve more than motion-oriented tasks.

\noindent\textbf{Attention Mechanism} explicitly calculates the correlation between two features \cite{bahdanau2015neural, lin2017structured}, and has been widely used in a variety of fields. For machine translation, the Transformer architecture first introduced by \citet{vaswani2017attention}, utilizes multi-head self-attention that captures diverse aspects in the input features \cite{voita2019analyzing}. For video QA, \citet{kim2018multimodal, li2019beyond} use self and guided-attention to encode temporal dynamics in video and ground them in the question. For multi-modal alignment, \citet{tsai2019multimodal} apply the Transformer to merge cross-modal time series between vision, language, and audio features. We utilize the attention mechanism to capture various relations between appearance and motion and to aggregate them.

\section{Model}
In this section, we introduce a detailed description of our MASN network. First, we explain how to obtain appearance and motion features in Section \ref{section:rep}. Then, we describe the Appearance and Motion modules, which encode visual features and combine them with the question in Section \ref{section:moapprmodule}. Finally, the Motion-Appearance Fusion module modulates the amount of motion and appearance information utilized and integrates them based on question context.

\subsection{Visual and Linguistic Representation}                \label{section:rep}
We first extract appearance and motion features from the video frames. For the appearance representation, we use ResNet \cite{he2016deep} pre-trained on an object and its attribute classification task as a feature extractor. For the motion representation, we use I3D \cite{carreira2017quo} pre-trained on the action classification task. We obtain local features representing object-level information without background noise and global features representing each frame's context for both appearance and motion features.

\paragraph{Appearance Representation.}
For local features, given a video containing $T$ frames, we obtain $N$ objects from each frame using Faster R-CNN \cite{ren2016faster} that applies RoIAlign to extract the region of interest from ResNet's convolutional layer. We denote the appearance-object set as $\mathcal{R}^{a} = \{ \mathbf{o}_{t,n}^{a}, \mathbf{b}_{t,n} \}_{t=1, n=1}^{t=T, n=N}$, where $\mathbf{o}$, $\mathbf{b}$ indicate object feature and bounding box location, respectively. Therefore, there are $K = N\times T$ objects in a single video. Following previous works, we extract the feature map from ResNet-152's \emph{Conv5} layer and apply a linear projection \cite{jiang2020divide, huang2020location}. We denote global features as $\mathbf{v}_{global}^{a} \in \mathbb{R}^{T \times d}$, where $d$ is the size of the hidden dimension.

\paragraph{Motion Representation.}
We obtain a feature map from the last convolutional layer in I3D \cite{carreira2017quo} whose dimension is (time, width, height, feature) = ($\left \lfloor \frac{t}{8} \right \rfloor$, 7, 7, 2048). That is, each set of 8 frames is represented as a single feature map with dimension $7 \times 7 \times 2048$. For local features, we apply RoIAlign \cite{he2017mask} on the feature map using object bounding box location $\mathbf{b}$. We define the motion-object set as $\mathcal{R}^{m} = \{ \mathbf{o}_{t,n}^{m}, \mathbf{b}_{t,n} \}_{t=1, n=1}^{t=T, n=N}$. We apply average pooling in the feature map and linear projection to obtain global features $\mathbf{v}_{global}^{m} \in \mathbb{R}^{T \times d}$.

\paragraph{Location Encoding.}
To reason about relations between objects as in Section \ref{section:moapprmodule}, it is required to consider each object's spatial and temporal location. As appearance and motion features share identical operations until the Motion-Appearance Fusion module, we combine superscript $a$ and $m$ for simplicity. Following L-GCN \cite{huang2020location}, we add a location encoding and define local features as:
\begin{equation}
    \mathbf{v}_{local}^{a/m} = \mathrm{FFN}([\mathbf{o}^{a/m}; \mathbf{d}^{s}; \mathbf{d}^{t}])
\end{equation}
where $\mathbf{d}^{s} = \mathrm{FFN}(\mathbf{b})$ and $\mathbf{d}^{t}$ is obtained by position encoding according to each frame's index. Here $\mathbf{o}^{a/m}$ denotes the object features mentioned above while $\mathrm{FFN}$ denotes a feed-forward network. Analogous to local features, position encoding information $\mathbf{d}^{t}$ is added to global features as well. We then concatenate object features with global features to reflect the frame-level context in objects and obtain the visual representation $\mathbf{v}^{a/m} \in \mathbb{R}^{K \times d}$:
\begin{equation}
\label{eq:locenc}
    \mathbf{v}^{a/m} = \mathrm{FFN}([\mathbf{v}_{local}^{a/m}; \mathbf{v}_{global}^{a/m}])
\end{equation}

\paragraph{Linguistic Representation.}
We apply the pre-trained GloVe to convert each question word into a 300-dimensional vector, following previous work \cite{jang2017tgif}. To represent contextual information in a sentence, we feed the word representations into a bidirectional LSTM (bi-LSTM). Word-level features and last hidden units from the bi-LSTM are denoted by $F^{q} \in \mathbb{R}^{L \times d}$, and $\mathbf{q} \in \mathbb{R}^{d}$ respectively. $L$ denotes the number of words in a question.

\subsection{Motion and Appearance Module}
\label{section:moapprmodule}

In this section, we explain the modules generating high-level visual representations and integrate them with linguistic representations. Each module consists of (1) an \textbf{Object Graph}: spatio-temporal reasoning between object-level visual features, and (2) \textbf{VQ interaction}: calculating correlations between objects and words and obtaining cross-modal feature embeddings. Since the modules share the same architecture, we describe each module's components only once with a shared superscript to avoid redundancy.

\subsubsection{Object Graph Construction}
In this section, we define object graphs $\mathcal{G}^{a/m} = (\mathcal{V}^{a/m}, \mathcal{E}^{a/m})$ to capture spatio-temporal relations between objects. $\mathcal{V}$, $\mathcal{E}$ denotes the node and edge set of the graph. As equation \ref{eq:locenc} provides visual features $\mathbf{v}^{a/m}$, we define these as the graph input $X^{a/m} \in \mathbb{R}^{K\times d}$. We denote the graph as $\mathcal{G}^{a/m}$. The nodes of graph $\mathcal{G}^{a/m}$ are given by $\mathnormal{v}_{i}^{a/m} \in X^{a/m}$, and edges are given by $(\mathnormal{v}_{i}^{a/m}, \mathnormal{v}_{j}^{a/m})$, representing a relationship between the two nodes.
Given the constructed graph $\mathcal{G}$, we perform graph convolution \cite{kipf2016semi} to obtain the relation-aware object features. We obtain the similarity scores of nodes by calculating the dot-product after projecting input features to the interaction space and define the adjacency matrix $A^{a/m} \in \mathbb{R}^{K \times K}$ as follows:
\begin{equation}
     A^{a/m} = \mathrm{softmax}((X^{a/m}W_{1})(X^{a/m}W_{2})^\top)
\end{equation}
We denote the two-layer graph convolution on input $X$ with adjacency matrix $A$ as:
\begin{equation}
\begin{gathered}
\label{eq:GCN}
    \mathrm{GCN}(X;A) = \mathrm{ReLU}(A\;\mathrm{ReLU}(AXW_{3})\;W_{4}) \\
    F = \mathrm{LayerNorm}(X + \mathrm{GCN}(X;A))
\end{gathered}
\end{equation}
We omit superscripts in the graph convolution equation for simplicity. We add a skip connection for residual learning between self-information $X$ and smoothed-information with neighbor objects.

\subsubsection{Vision-question (VQ) Interaction} \label{section:vqinteract}
We compute both appearance-question and motion-question interaction to obtain correlations between language and each of the visual features. As we encode visual feature $F^{a/m}$ and question feature $F^{q}$ in Equation \ref{eq:GCN} and Section \ref{section:rep}, we calculate every pair of relations between two modalities using the bilinear operation introduced in BAN \cite{kim2018bilinear} as follows:
\begin{equation}
    H_{i} = \mathds{1} \cdot \mathrm{BAN}_{i}(H_{i-1}, V; \mathcal{A}_{i})^\top + H_{i-1}
\end{equation}
where $H_{0} = F^{q}$, $\mathds{1} \in \mathbb{R}^{L}$, $1 \leq i \leq g$ and $\mathcal{A}$ denotes the attention map. $F^{a/m}$ is substituted for $V$ respectively in our method. In the equation above, calculating the result $\mathrm{BAN}(H, V; \mathcal{A}) \in \mathbb{R}^{d}$ and adding it to the $H$ is repeated in $g$ times. Afterwards, $H$ represents the combined visual and language features in the question space incorporating diverse aspects from the two modalities \cite{yang2016stacked}.

\subsection{Motion-Appearance Fusion}
In this section, we introduce the Motion-Appearance Fusion module which is our key contribution. Depending on what the question ultimately asks about, the model is supposed to decide which features are more relevant among appearance and motion information, or a combination of both. To do this, we produce appearance-centered, motion-centered, and all-mixed features and aggregate them depending on question context. Based on the previous step, we obtain cross-modal combined features $H^{a}$ and $H^{m}$ in terms of appearance and motion. We concatenate these two matrices and define $U$ as:
\begin{equation}
\label{eq:attinp}
    U = \begin{bmatrix}
    H^{a}\\ 
    H^{m}
    \end{bmatrix}, \; U \in \mathbb{R}^{2L \times d}
\end{equation}

\begin{figure}[t]
\centering
    \includegraphics[width=1.0\columnwidth]{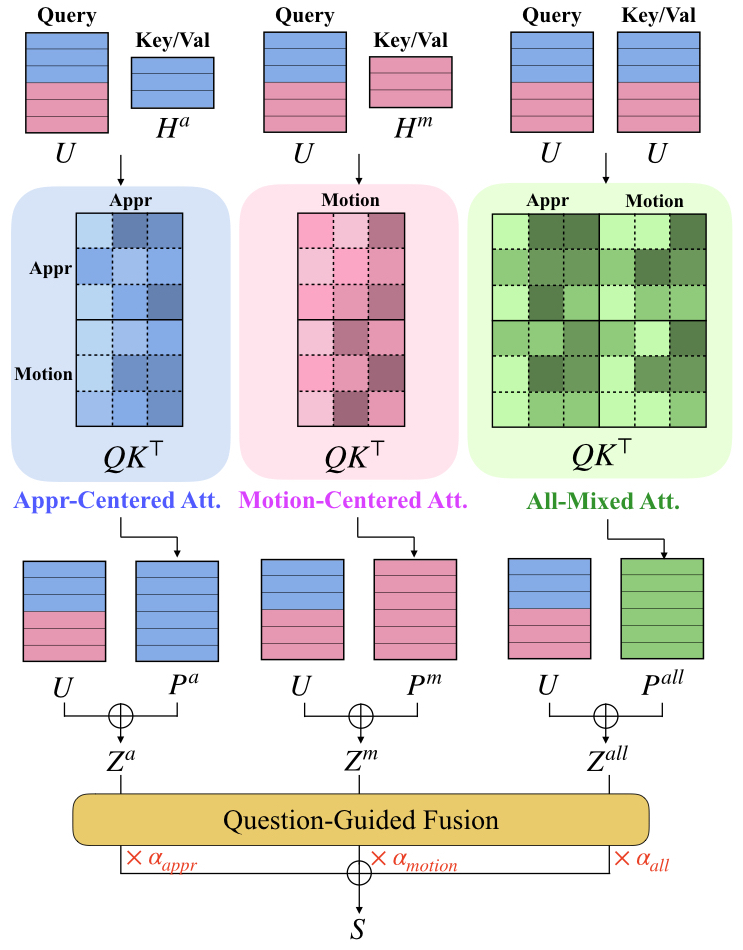}
\caption{Motion-Appearance Fusion module. The blue-colored elements in a matrix denote appearance-question, and the pink ones indicate motion-question combined features. Matrices above $QK^\top$ represent an attention score maps from each kind of attention. The final output $S$ in the figure is the weighted-sum matrix of all three attended features.}
\label{fig:model2}
\end{figure}

\paragraph{Motion-Appearance-centered Attention.}
We first define regular scaled dot-product attention to attend features to diverse aspects:
\begin{equation}
\label{eq:scaledotatt}
    \mathrm{Attention}(Q, K, V) = \mathrm{softmax}(\frac{QK^\top}{\sqrt{d_{k}}})V
\end{equation}
where $Q$, $K$, $V$ denotes the query, key, and value, respectively. To obtain motion-centered, appearance-centered and mixed attention, we substitute $U$ with the query, and $H^{a}, H^{m}, U$ with the key and value in the equation \ref{eq:scaledotatt} as:
\begin{equation}
\label{eq:3att}
\begin{gathered}
    P^{a} = \mathrm{Attention}(U, ~H^{a}, ~H^{a}) \\
    P^{m} = \mathrm{Attention}(U, H^{m}, H^{m}) \\
    P^{all} = \mathrm{Attention}(U,~U,~U) \\
    Z^{a/m/all} = \mathrm{LayerNorm}(P^{a/m/all} + U)
\end{gathered}
\end{equation}
where $P \in \mathbb{R}^{2L \times d}$ and $Z \in \mathbb{R}^{2L \times d}$. 

As in the first line of the equation \ref{eq:3att}, we add projected appearance features $P^{a}$ on each appearance and motion feature to obtain $Z^{a}$, since the matrix $U$ is the concatenation of $H^{a}$ and $H^{m}$. Therefore, we argue that $Z^{a}$ contains appearance-centered information. Similarly, $Z^{m/all}$ contains motion-centered and all-mixed features, respectively. We argue that the Motion-Appearance-centered attention fuses appearance and motion features in various proportions and these three matrices work like multi-head attention sharing the task of capturing diverse information, and become synergistic when combined.

\paragraph{Question-Guided Fusion.}
For question-guided fusion, we first define $\mathbf{z}^{a/m/all}$ as the sum of matrix $Z^{a/m/all} \in \mathbb{R}^{2L \times d}$ over sequence length $2L$. We obtain attention scores between each $\mathbf{z}^{a/m/all}$ and question context vector $\mathbf{q}$:
\begin{equation}
\label{eq:qscore}
\begin{gathered}
    \alpha^{a/m/all} = \mathrm{softmax}(\frac{\mathbf{q}(\mathbf{z}^{a/m/all})^\top}{\sqrt{d_{z}}}) \\
\end{gathered}
\end{equation}
where $\mathbf{q}$ denotes the last hidden vector. The attention score $\alpha^{a/m/all}$ can be interpreted as the importance of each matrix $Z$ based on question context. We obtain the question-guided fusion matrix $O$ as:
\begin{equation}
\begin{gathered}
    S = \alpha^{a}Z^{a} + \alpha^{m}Z^{m} + \alpha^{all}Z^{all} \\
    O = \mathrm{LayerNorm}(S + \mathrm{FFN}(S))
\end{gathered}
\end{equation}
where $O \in \mathbb{R}^{2L \times d}$ is obtained by linear transformation and a residual connection after weighted sum. We aggregate information by attention over the sequence length of $O$:
\begin{equation}
\label{eq:attflat}
\begin{gathered}
    \beta_{i} = \mathrm{softmax}(\mathrm{FFN}(O_{i})) \\
    \mathbf{f} = \sum_{i=1}^{2L} \beta_{i}O_{i}
\end{gathered}
\end{equation}
The final output vector $\mathbf{f} \in \mathbb{R}^{d}$ is used for answer prediction.

\subsection{Answer Prediction and Loss Function}
The video QA task can be divided into counting, open-ended word, and multiple-choice tasks \cite{jang2017tgif}. Our method trains the model and predicts the answer based on the three tasks similar to previous work.

The counting task is formulated as a linear regression of the final output vector $\mathbf{f}$. We obtain the final answer by rounding the result and we minimize Mean Squared Error (MSE) loss.

The open-ended word task is essentially a classification task over the whole answer set. We calculate a classification score by applying a linear classifier and softmax function on the final output $\mathbf{f}$ and train the model by minimizing cross-entropy loss.

For the multiple-choice task, like in previous work \cite{jang2017tgif}, we attach an answer to the question and obtain $M$ candidates. Then, we obtain the score for each of the $M$ candidates by a linear transformation to the output vector $\mathbf{f}$. We minimize the hinge loss within every pair of candidates, $max(0, 1+s_{n}-s_{p})$, where $s_{n}$ and $s_{p}$ are scores from incorrect and correct answers respectively.

\section{Experiments}
In this section, we evaluate our proposed model on three Video QA datasets: TGIF-QA, MSVD-QA, and MSRVTT-QA. We first introduce each dataset and compare our results with the state-of-the-art methods. Then, we report ablation studies and include visualizations to show how each module in MASN works.

\subsection{Datasets}

\textbf{TGIF-QA} \cite{jang2017tgif} is a large-scale dataset that consists of 165K QA pairs collected from 72K animated GIFs. The length of video clips is very short, in general. TGIF-QA consists of four types of tasks: Count, Action, State transition (Trans.), and FrameQA. Count is an open-ended question to count how many times an action repeats. Action is a task to find action repeated at certain times, and Transition aims to identify a state transition over time. Both types are multiple-choice tasks. Lastly, FrameQA is an open-ended question that can be solved from just one frame, similar to image QA. \\

\noindent\textbf{MSVD-QA \& MSRVTT-QA} \cite{xu2017video} are automatically generated from video descriptions. It consists of 1,970 video clips and 50K and 243K QA pairs, respectively. The average video lengths are 10 seconds and 15 seconds respectively. Questions belong to five types: what, who, how, when, and where. The task is open-ended with a pre-defined answer sets of size 1,000 and 4,000, respectively.


\renewcommand{\ULdepth}{1pt} 
\begin{table}[t]
\centering
\normalem
\begin{tabular}{l|cccc}
\Xhline{2\arrayrulewidth}
Methods & Count & Action & Trans. & FrameQA\\
\Xhline{2\arrayrulewidth}
ST-VQA & 4.28 & 60.8 & 67.1 & 49.3 \\
\hline
Co-Mem & 4.10 & 68.2 & 74.3 & 51.5 \\
PSAC & 4.27 & 70.4 & 76.9 & 55.7 \\
STA & 4.25 & 72.3 & 79.0 & 56.6 \\
HME & 4.02 & 73.9 & 77.8 & 53.8 \\
\hline
HGA & 4.09 & 75.4 & 81.0 & 55.1 \\
L-GCN & 3.95 & 74.3 & 81.1 & 56.3 \\
QueST & 4.19 & \uline{75.9} & 81.0 & \textbf{59.7} \\
HCRN & \uline{3.82} & 75.0 & \uline{81.4} & 55.9 \\
\hline
MASN & \textbf{3.75} & \textbf{84.4} & \textbf{87.4} & \uline{59.5} \\
\Xhline{2\arrayrulewidth}
\end{tabular}
\caption{\label{table:tgif}
State-of-the-art comparison on the TGIF-QA dataset. Mean $\ell_{2}$ loss for Count, and accuracy (\%) for others. Best results in bold, underlined results denote the second best.
}
\end{table}


\begin{table}[t]
\centering
\normalem
\begin{tabular}{l|cc}
\Xhline{2\arrayrulewidth}
Methods & MSVD-QA & MSRVTT-QA \\ 
\Xhline{2\arrayrulewidth}
ST-VQA & 31.3 & 30.9 \\
GRA & 32.0 & 32.5 \\
Co-Mem & 31.7 & 32.0 \\
HME & 33.7 & 33.0 \\
\hline
HGA & 34.7 & \uline{35.5} \\
QuesT & \uline{36.1} & 34.6 \\
HCRN & \uline{36.1} & \textbf{35.6} \\
\hline
MASN & \textbf{38.0} & 35.2 \\
\Xhline{2\arrayrulewidth}
\end{tabular}
\caption{\label{table:msqa} State-of-the-art comparison on the MSVD-QA and MSRVTT-QA datasets. All values represent accuracy (\%). Best results in bold, underlined results denote the second best.}
\end{table}

\begin{table*}[t]
\normalem
\centering
\begin{tabular}{>{\centering\arraybackslash}m{10em}|>{\centering\arraybackslash}m{8em}|>{\centering\arraybackslash}m{3.5em}>{\centering\arraybackslash}m{3.5em}>{\centering\arraybackslash}m{3.5em}>{\centering\arraybackslash}m{3.5em}}
\Xhline{2\arrayrulewidth}
\multicolumn{2}{c|}{Methods} & Count & Action & Trans. & FrameQA\\
\Xhline{2\arrayrulewidth}
\multicolumn{2}{c|}{Appr. Module} & 3.94 & 82.9 & 86.2 & 58.6 \\
\multicolumn{2}{c|}{Motion Module} & 3.84 & 82.5 & 86.2 & 51.0 \\
\multicolumn{2}{c|}{Appr. Module + Motion Module} & 3.82 & 83.4 & 86.8 & 58.6 \\
\multicolumn{2}{c|}{Appr. Module + Motion Module + Fusion (Ours)} & \textbf{3.75} & \textbf{84.4} & \uline{87.4} & \textbf{59.5} \\
\hline\hline
& Appr. & 3.78 & 82.8 & 86.3 & 58.9 \\ \cline{2-6}
Single-Att. Fusion & Motion & 3.79 & 83.1 & 87.0 & 59.1 \\ \cline{2-6}
& All & 3.78 & 83.6 & \uline{87.4} & \uline{59.3} \\
\hline
& Appr. + Motion & \uline{3.77} & 83.6 & \uline{87.4} & 59.2 \\ \cline{2-6}
Dual-Att. Fusion & Appr. + All & \uline{3.77} & 83.6 & \textbf{87.5} & 59.0 \\ \cline{2-6}
& Motion + All & 3.80 & \uline{84.1} & 86.5 & 59.1 \\
\Xhline{2\arrayrulewidth}
\end{tabular}
\caption{\label{table:ablation}
Ablation study on the TGIF-QA dataset. Mean $\ell_{2}$ loss for Count, and accuracy (\%) for others. Appr. and Att. stand for Appearance and Attention. Best results in bold, underlined results denote the second best.
}
\end{table*}

\subsection{Implementation Details}
We first extract frames with 6 fps for all datasets. In the case of \textbf{appearance features}, we sample 1 frame out of 4 to avoid information redundancy. We apply Faster R-CNN \cite{ren2016faster} pre-trained on Visual Genome \cite{krishna2017visual} to obtain local features. The number of extracted objects is $N=10$. For global features, we use ResNet-152 pre-trained on ImageNet \cite{deng2009imagenet}. In the the case of \textbf{motion features}, we apply I3D pre-trained on the Kinetics action recognition dataset \cite{kay2017kinetics}. For the input of I3D, we concatenate a set of 8 frames around the sampled frame mentioned above. In terms of training details, we employ Adam optimizer with learning rate as $10^{-4}$. The number of BAN glimpse $g$ is 4. We set the batch size as 32 for the Count and FrameQA tasks and 16 for Action and Trans. tasks.

\subsection{Comparison with State-of-the-arts}

We compare MASN with state-of-the-art (SoTA) models on the aforementioned datasets.
\paragraph{TGIF-QA.}
Compared with ST-VQA \cite{jang2017tgif}, Co-Mem \cite{gao2018motion}, PSAC \cite{li2019beyond}, STA \cite{gao2019structured}, HME \cite{fan2019heterogeneous}, and recent SoTA models: HGA, L-GCN, QueST, HCRN \cite{jiang2020reasoning, huang2020location, jiang2020divide, le2020hierarchical}, MASN shows the best results for three tasks: Count, Trans., and Action, outperforming the baseline methods by a large margin as shown in Table \ref{table:tgif}. In the case of FrameQA, the performance is similar to QueST. However, considering that there exists some tradeoff between the performance of Count and FrameQA since Count focuses on identifying temporal information and FrameQA focuses on spatial information, MASN shows the best overall performance on the entire task. 

\paragraph{MSVD-QA \& MSRVTT-QA.}
As shown in Table \ref{table:msqa}, MASN outperforms the best baseline methods, QuesT and HCRN by approximately 2\% on MSVD-QA, and shows competitive results on MSRVTT-QA. Since these datasets are composed of wh-questions, such as what or who, the question sets seemingly resemble FrameQA in TGIF-QA, as they tend to focus on spatial appearance features. This means that MASN is able to capture spatial details well based on the spatiotemporally mixed features.


\begin{figure*}[t]
\centering
    \subfigure{\includegraphics[width=16cm]{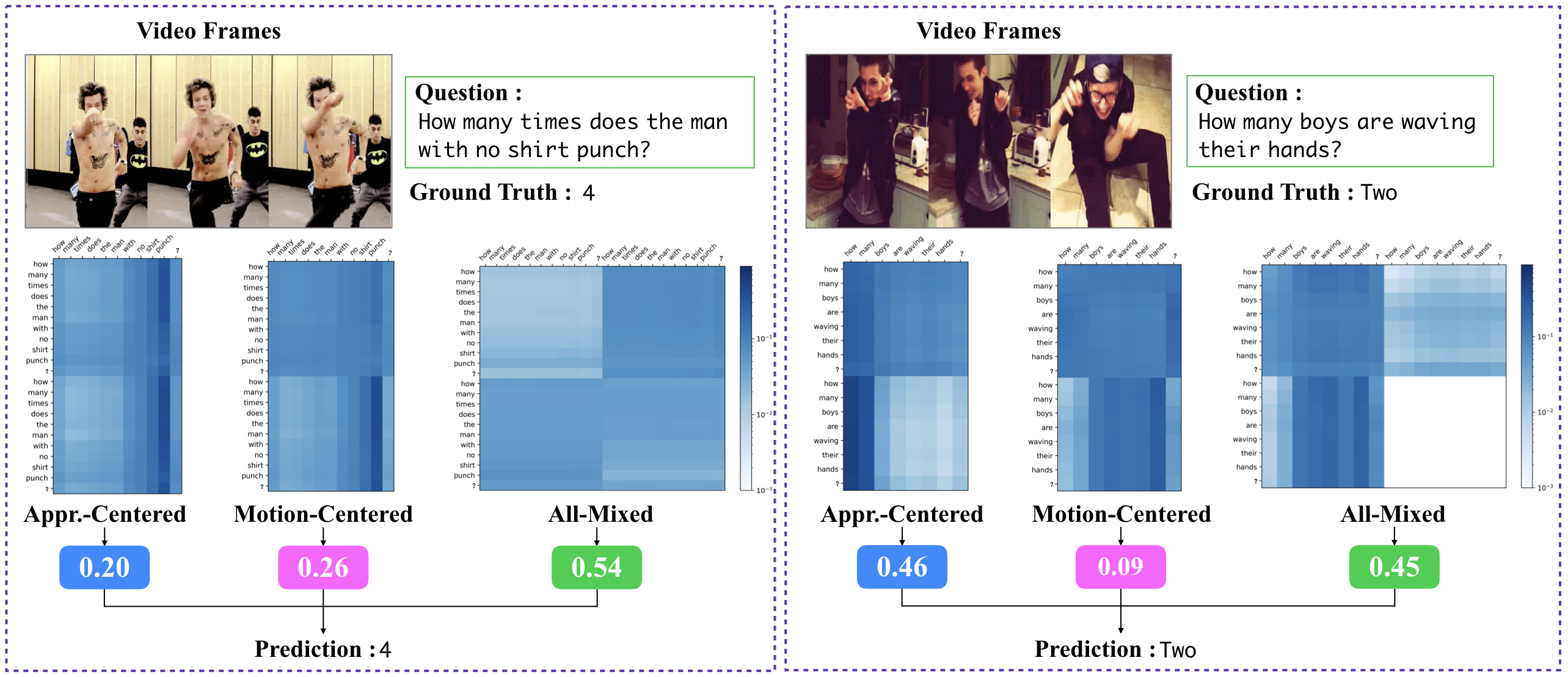}}
    \subfigure{\includegraphics[width=16cm]{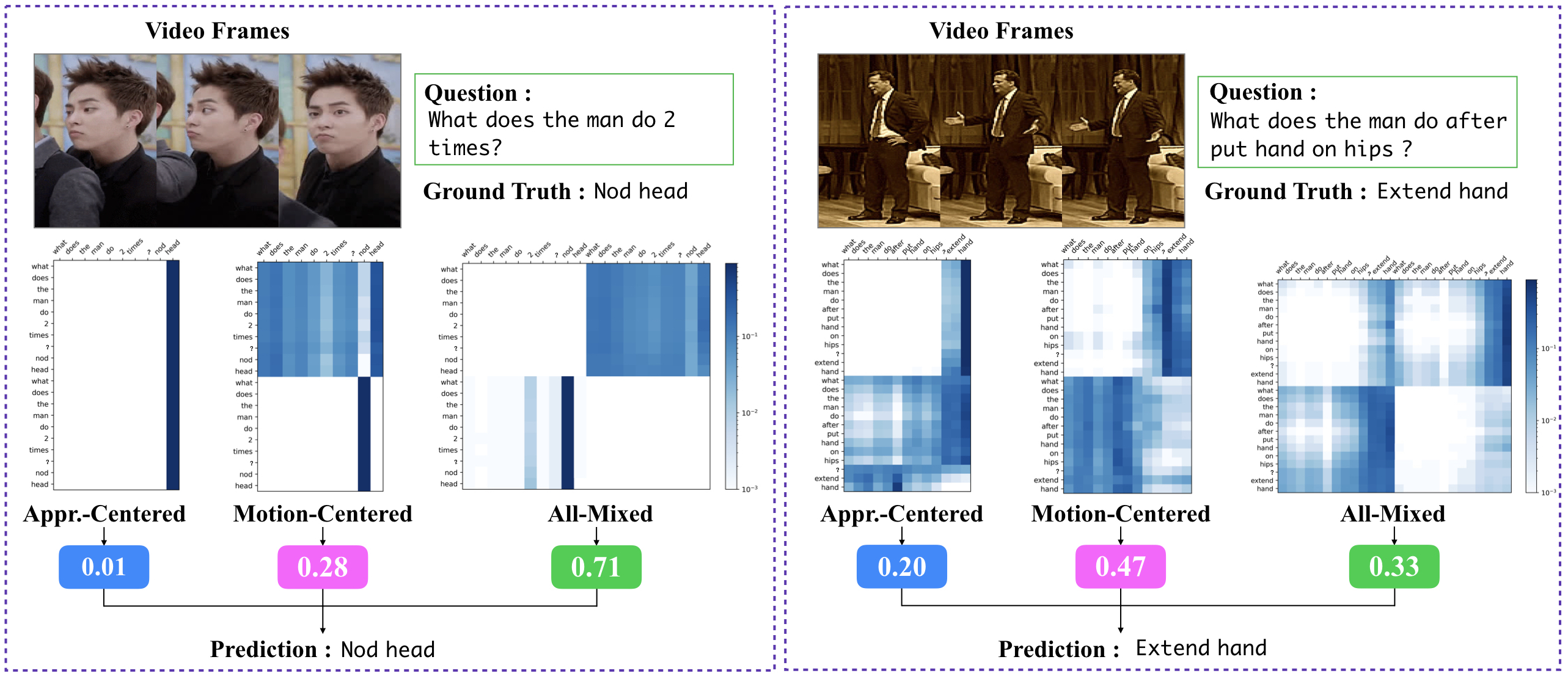}}
\caption{Qualitative results on TGIF-QA dataset. From the left, Count and FrameQA are shown in 1st row and Action, Trans. in 2nd row. Each visualized attention map is log-scaled. Scores below attention maps represent $\alpha$ from the equation \ref{eq:qscore}.}
\label{fig:qualitative}
\end{figure*}

\subsection{Ablation Study}

\paragraph{Analyzing the impact of motion module and appearance module.}
We investigate the effect of each module as seen in Figure \ref{fig:model1}. In Table \ref{table:ablation}, the $1^{st}$ and $2^{nd}$ row represent the result of using only the Appearance and Motion module, respectively. The $3^{rd}$ row shows the result of just concatenating appearance and motion features from each module and flattening them, by substituting the input $X$ for $O$ in equation \ref{eq:attflat}. Most existing SOTA models utilize only ResNet features for spatio-temporal reasoning based on the difference of vectors over time. Using only the Appearance module is similar to most of these existing methods, which can catch spatio-temporal relations relatively well. On the other hand, we found that the accuracy on FrameQA when only using the Motion module is about 7\% lower than when using the Appearance module. This means the Motion module is limited in its ability to capture the appearance details. However, comparing the $1^{st}$ and $3^{rd}$ row in Table 3, the performance in the Action and Trans. tasks increase consistently when the Motion module is added compared to using only the Appearance module. This indicates that the Motion module is a meaningful addition. Lastly, compared to the $1^{st}$, $2^{nd}$ and $3^{rd}$ row, when integrating the output from both modules there is a further overall performance improvement. This indicates a synergistic effect occurs when integrating both the appearance and motion feature after obtaining them as high-level features.

\paragraph{Analyzing the impact of fusion module.}
We show ablation studies inside the fusion module represented in Table \ref{table:ablation}. The $4^{th}$ row indicates the performance of our proposed MASN architecture. The results in the `Single-Attention Fusion' row use only one type of attention among appearance-centered, motion-centered, and all-mixed as seen in equation \ref{eq:3att}. The results in the `Dual-Attention Fusion' row utilize two among the three types of attention mentioned above. Due to the nature of video, when a question such as “How many times does the man in the white shirt put his hand on the head?” is given, the model is supposed to find the motion information “put” while catching the appearance information “man in white shirt” or “hand on head”, and finally mixing them in different proportions depending on the context of question. Comparing the result of the $3^{rd}$ (without fusion) row and MASN first, MASN shows better performance across tasks. This means mixing appearance and motion features in various proportions using the Motion-Appearance-centered Fusion module and computing the weighted fusion via the Question-Guided Fusion module contributes to the performance. When comparing the general performance with the number of attention types in fusion module, using single, dual, and triple attention (MASN) shows increasingly better performance in the same order. This indicates that focusing on different aspects and integrating each attended feature performs better than calculating attention at once. Additionally, comparing the result of using only appearance or motion-centered attention in `Single' with both of them in `Dual', we find that using both features shows better performance, which means they play complementary roles for each other. Similarly, we argue the reason for the performance increase in FrameQA in the ‘Motion’ row of ‘Single-Att. Fusion’ is due to the fact that the model can find relevant appearance information better based on motion information.


\subsection{Qualitative Results}
We give examples of each attention score matrix from Motion-Appearance Fusion module in Figure \ref{fig:qualitative}. We draw two conclusions from the Figure: (1) each attention map catches different relations similarly to multi-head attention, (2) each attention map is used to a different extend depending on the type of task. For example, in FrameQA, the appearance-centered's attention map captures which appearance trait to find focusing on `how many'. On the other hand, the motion-centered's and all-mixed's attention map attend on `waving' or `hands' to catch motion-related information. In Action, similar to FrameQA, the appearance-centered's attention map attends on `head' which is the object of action, while the motion-centered's attention map catch `nod' which is related to movement. However, in the case of the Count task, the two attention weights are not as sparse as scores in the other tasks. We think this dense attention map causes the inconsistency in the performance increase between Count task and Action and Trans. task, although questions for all of these three tasks ask for motion information.

\section{Conclusion}
In this paper, we proposed a Motion-Appearance Synergistic Networks to fuse and create a synergy between motion and appearance features. Through the Motion and Appearance modules, MASN manages to find motion and appearance clues to solve the question, while modulating the amount of information used of each type through the Fusion module. Experimental results on three benchmark datasets show the effectiveness of our proposed MASN architecture compared to other models.

\paragraph{Acknowledgement}
The authors would like to thank Ho-Joon Song, Yu-Jung Heo, Bjorn Bebensee, Seonil Son, Kyoung-Woon On, Seongho Choi and Woo-Suk Choi for helpful comments and editing. This work was partly supported by the Institute of Information \& Communications Technology Planning \& Evaluation (2015-0-00310-SW.StarLab/25\%, 2017-0-01772-VTT/25\%, 2018-0-00622-RMI/25\%, 2019-0-01371-BabyMind/25\%) grant funded by the Korean government.

\bibliographystyle{acl_natbib}
\bibliography{anthology,acl2021}


\end{document}